\pdfoutput=1
\documentclass[11pt,a4paper]{article}
\usepackage[hyperref]{acl2021}
\usepackage{times}
\usepackage{latexsym}

\usepackage{microtype}

\usepackage{algorithm}
\usepackage[noend]{algpseudocode}
\usepackage{amsmath}
\usepackage{amssymb}
\usepackage{booktabs}
\usepackage{mathtools}
\usepackage{pgfplots}
\usepackage{stmaryrd}
\usepackage{subfig}
\usepackage{tabularx}
\usepackage{tikz}

\DeclareMathOperator{\softmax}{softmax}
\DeclareMathOperator{\relu}{ReLU}

\usetikzlibrary{arrows.meta,calc}
\tikzset{>=latex}

\pgfplotsset{compat=newest}
\pgfplotsset{every axis/.append style={xticklabel style={font=\small},
yticklabel style={font=\small}}}
\usepgfplotslibrary{colorbrewer}

\aclfinalcopy 

\newcommand{\rf}{Ref}
\newcommand{\cai}{Cai \& Lam '20}
\newcommand{\cais}{CL20}
\newcommand{\chrf}{\textsc{chrF++}}
\newcommand{\tdp}{\textsc{TDP}}
\newcommand{\tdpm}{\textsc{TDP-M}}
\newcommand{\tree}{\textsc{Tree}}
\newcommand{\ldc}{LDC2017T10}
\newcommand{\mol}{MolHIV}

\title{Latent Tree Decomposition Parsers for AMR-to-Text Generation}
\author{Lisa Jin \and Daniel Gildea\\
  Department of Computer Science\\
  University of Rochester\\
Rochester, NY 14627}

\date{}

\begin{document}
\maketitle
\begin{abstract}
  Graph encoders in AMR-to-text generation models often rely on neighborhood convolutions or global vertex attention. While these approaches apply to general graphs, AMRs may be amenable to encoders that target their tree-like structure. By clustering edges into a hierarchy, a tree decomposition summarizes graph structure. Our model encodes a derivation forest of tree decompositions and extracts an expected tree. From tree node embeddings, it builds graph edge features used in vertex attention of the graph encoder. Encoding TD forests instead of shortest-pairwise paths in a self-attentive baseline raises BLEU by 0.7 and \chrf{} by 0.3. The forest encoder also surpasses a convolutional baseline for molecular property prediction by 1.92\% ROC-AUC.
\end{abstract}

\section{Introduction}
Due to the prevalence of structured data such as graphs or tables, models that generate text from these formats are highly practical. The models must produce fluent sequences without deviating from the input semantics. A major obstacle for this task is the disconnect between the structure and order of source and target elements.

Abstract Meaning Representation \cite{banarescu2013abstract} is a semantic graph that captures sentence-level propositional meaning. It has labeled vertices and edges called concepts and relations, respectively. AMRs are typically tree-like and rooted at one concept from which relations point out. AMRs that are not strictly trees have reentrancies, or vertices with multiple parents. Besides AMR's denser structure, its underspecified semantic labels also challenge text generation.

Recent AMR-to-text models \cite{zhu2019modeling,cai2020graph} favor encoders that model global vertex dependencies. Such models often adopt self-attention from the sequence encoder of the Transformer \cite{vaswani2017attention}. Unlike graph convolutional networks (GCNs), self-attention updates each vertex as a function of the entire graph. In place of the recurrent updates found in GCN models \cite{beck2018graph,song2017amr}, these graph encoders apply stacked layers of attention. Thus, vertices in self-attentive graph encoders enjoy constant-time communication, regardless of their distance.

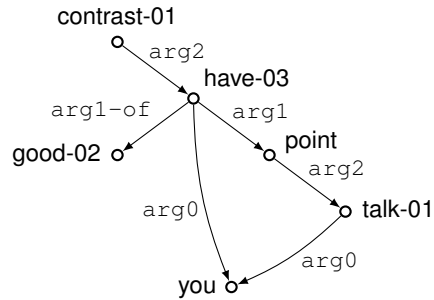
\begin{figure}
  \centering
    \begin{tikzpicture}
      \begin{scope}[font=\footnotesize\sffamily,
        every node/.style={circle,thick,draw,inner sep=0.125em}]
        \node (c) at (0,0) [label={[inner sep=0,shift={(0,-.55)}]contrast-01}] {};
        \node (g) at (0,-1.5) [label={[inner sep=0,shift={(-.8,-.7)}]good-02}] {};
        \node (h) at (1,-.75) [label={[inner sep=0,shift={(.7,-.4)}]have-03}] {};
        \node (p) at (2,-1.5) [label={[inner sep=0,shift={(.55,-.3)}]point}] {};
        \node (t) at (3,-2.25) [label={[inner sep=0,shift={(.7,-.55)}]talk-01}] {};

        \node (y) at (1.5,-3.25) [label={[inner sep=0,shift={(-.45,-.4)}]you}] {};
      \end{scope}
      \begin{scope}[font=\footnotesize\ttfamily,>=latex,auto,
        every node/.style={fill=none,circle,inner sep=0},
        every edge/.style={draw=black,thin}]
        \path [->] (c) edge node [shift={(0,-.05)}] {arg2} (h);
        \path [->] (h) edge node [shift={(-1.2,.7)}] {arg1-of} (g);
        \path [->] (h) edge node [shift={(.1,-.1)}] {arg1} (p);
        \path [->] (p) edge node [shift={(0.1,-.1)}] {arg2} (t);
        \path [->,bend right=10] (h) edge node [shift={(-.7,-.5)}] {arg0} (y);
        \path [->,bend left=10] (t) edge node [shift={(.2,.2)}] {arg0} (y);
      \end{scope}
    \end{tikzpicture}
  \caption{AMR for the sentence, ``But good you have your talking points!''}
  \label{fig:sample-amr}
\end{figure}

Despite their flexibility, self-attentive encoders ignore graph structure. One way to refine self-attention is to mimic GCNs by masking out non-adjacent vertices \cite{zhu2019modeling,wang2020amr}. \citet{cai2020graph} instead bias embeddings of each pair of query-key vertices on the shortest path between them. The first approach faces the drawbacks of GCNs, while the second only partially captures per-vertex graph context.

Instead of encoding graph topology explicitly, we aim to close the \textit{structural gap} between graph and string \cite{zhao2020bridging}. We train a latent parser that extracts tree decompositions, which cluster edges of the input graph into a hierarchy. Whereas these trees can be used to efficiently marginalize over variables of a graphical model, we use them to bridge graph and string. Specifically, the trees simplify graph structure and ease alignment between predicted tokens and vertices.

To further motivate the use of tree decompositions for graph-to-string models, we note their parallels to context-free graph grammars. Bags of a tree decomposition map to states of a derivation tree in a hyperedge replacement grammar \cite{lautemann1988decomposition}. In effect, the same algorithm that applies the graph grammar can be used to parse a forest of tree decompositions. \citet{jones2013modeling} present HRGs as analogous to context-free string and tree grammars; they further show how HRGs can be drawn from tree decompositions. They do not, however, apply the HRGs to concrete tasks.

Tree decompositions provide benefits as (i) a structural intermediate between graph and string and (ii) a model of AMR graph languages. This work integrates a tree decomposition parser into the graph encoder of an AMR-to-text model. Our encoder uses the node embeddings of an expected tree to bias self-attention following \citet{cai2020graph}. While they use shortest paths to learn edge embeddings between all vertex pairs, we only update existing edges. By parsing tree decompositions, our model performs exact inference over the full graph instead of relying on local path features.

We further experiment on the molecular chemistry \mol{} dataset \cite{hu2020ogb} to test the parser on graph classification. Previous work leveraged tree-structured scaffolds over molecular graphs to help constrain the chemical validity of a generative model \cite{jin2018junction}. The bags of our tree decompositions are domain invariant and thus do not necessarily map to valid chemical substructures. The promising results of our tree decomposition parser in this new setting support its flexibility.

\section{Tree Decomposition}
We now introduce tree decomposition and how it applies to graphs of a certain treewidth. We then expose parallels between tree decomposition and hyperedge replacement grammars. Lastly, we review an $\mathcal{O}(n^{k + 1})$ algorithm that parses $k$-width tree decompositions from a graph.

\subsection{Definition}
\label{sec:td-def}
A tree decomposition reduces a graph to a hierarchy of subgraphs. Given graph $G = (V, E)$ it is denoted by $T = (I, F)$, where $\{X_i\}_{i \in I}$ contains vertex sets, or bags, that are linked by arcs in $F$. Tree decompositions have the below properties.
\begin{enumerate}
  \item Vertex cover: The bags of $T$ cover all vertices in $V$, or $\bigcup_i X_i = V$.
  \item Edge cover: Each edge $(u, v) \in E$ is covered by some bag of $T$, or $\exists i\colon u, v \in X_i$.
  \item Running intersection: All bags on the path from $X_i$ to $X_j$ in $T$ cover vertices in $X_i \cap X_j$.
\end{enumerate}
An optimal tree decomposition (in short TD) of a graph has minimal width, which relates to the largest bag: $\max_i |X_i| - 1$. A bag is capable of covering a clique among its vertices; graphs of greater complexity need larger bags to cover their edges. Thus, a graph's treewidth, or width of its optimal TD, helps characterize its complexity. For example, a tree lacks cliques of size greater than two and thus has treewidth 1. In contrast, an $n \times n$ grid graph has treewidth $n$; the optimal TD's bags chain together rows or columns of the grid.

As previously noted, TDs enable linear time inference in graphical models. Limiting TD width bounds the complexity of message passing among adjacent nodes. This improved efficiency via constant treewidth holds for a variety of NP-complete graph problems \cite{arnborg1991easy}.

\subsection{Relation to HRGs}
\label{sec:rel-hrg}
As HRG derivation trees and TDs have the same structure they can share a parser. We describe parsing TDs in terms of HRGs for clear intuition then define how the former maps onto the latter.

\begin{figure}
  \centering
  \begin{tikzpicture}[baseline={(c.base)}]
    \begin{scope}[every node/.style={inner sep=.1em},font=\small]
      \node (c) at (0,0) {$c$};
      \node (g) [below left of=c] {$g$};
      \node (h) [below right of=c] {$h$};
      \node (y) [below left of=h] {$y$};
      \node (p) [below right of=h] {$p$};
      \node (t) [below right of=y] {$t$};
    \end{scope}
    \draw [->] (c) -- (h);
    \draw [->] (h) -- (g);
    \draw [->] (h) -- (y);
    \draw [->] (h) -- (p);
    \draw [->] (p) -- (t);
    \draw [->] (t) -- (y);
  \end{tikzpicture}
  \qquad
  \begin{tikzpicture}[baseline={(a.base)},
    fork/.style={draw,rounded corners=.5em,->}]
    \begin{scope}[every node/.style={inner sep=.1em},font=\small,
      fnode/.style={inner sep=.2em,rounded corners=.1em,draw=teal!70,thick}]
      \node [fnode] (a) at (1,0) {$c,h$};
      \node [fnode] (c) at (0.4,-1.5) {$h,y,p$};
      \node [fnode] (d) at (1.6,-1.5) {$h,g$};
      \node [fnode] (e) at (0.4,-2.5) {$y,p,t$};
      \node [fnode] (f) at (-.25,0) {$c,h,g$};
    \end{scope}
    \path [fork] (c) |- ($(a)+(0,-.85)$) -| (a);
    \path [fork] (d) |- ($(a)+(0,-.85)$) -| (a);
    \path [->] (c.north) edge [bend left=15] (f);
    \path [draw,->] (e) -- (c);
  \end{tikzpicture}
  \label{fig:td-forest}
  \caption{Simplified AMR (left), forest fragment (right). Note that forest nodes $\{h,y,p\}$ and $\{y,p,t\}$ can be replaced by $\{h,y,t\}$ and $\{h,t,p\}$, respectively.}
\end{figure}
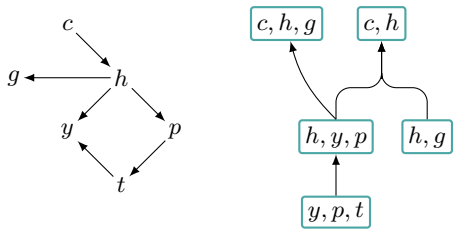

HRG is a context-free graph grammar that recursively replaces nonterminals called hyperedges with hypergraphs. An HRG is denoted by a tuple $\langle N, T, P, S\rangle$ of nonterminals $N$, terminals $T$, and start symbol $S \in N$. Productions $P$ take form $A \to R$ with $A \in N$ and $R \in N \cup T$. Hypergraph fragment $R$ includes a hypergraph $\langle V, E, \ell\rangle$ and external node set $X \in V^+$; the latter is an interface between the hypergraph and hyperedge $A$.

A TD can be formed by mapping the external node sets of an HRG derivation tree to bags. The derivation tree's nodes are HRG rules, which are linked by chained hyperedge replacement. Each node covers a subgraph of the input graph. When a node has children, this subgraph is split into smaller edge-disjoint subgraphs. Each child subgraph shares external nodes $X$ with its parent and exposes new edges that have not been seen in ancestor nodes. Thus, we can extract a TD by ignoring the nonterminals of each derivation tree node, keeping only the terminal vertices in $X$. Since a derivation tree rewrites all hyperedges by terminals, vertex and edge cover are satisfied. As external nodes $X$ exist in the parent node's hyperedge, running intersection is enforced as well.

\subsection{Parsing algorithm}
In order to extract TDs of at most width $k$ from an AMR, we employ an algorithm by \citet{lautemann1988decomposition}. The method recursively checks whether a given vertex set $S \subseteq V$ with $|S| \leq k + 1$ can serve as the subtree root of a TD for graph $G$.

\begin{algorithm}[ht]
  \caption{Recognition of $k$-width TD}
  \label{alg:decomp}
  \begin{algorithmic}[1]
  \Require $pairs \gets \{(S_i, C_{i\cdot})\}$ from graph $G$
    \Procedure{Decomp}{$S, C, k$}
    \If{$|\mathcal{N}_S(C) \cup C| \leq k + 1$} \Return \texttt{true} \EndIf
      \State $root \gets$ \texttt{false}
      \ForAll{$S^\prime \colon \mathcal{N}_S(C) \subset S^\prime \subseteq \mathcal{N}_S(C) \cup C$
      \\\hspace{5.85em}and $|S^\prime| \leq k + 1$}
      \State $ch\_root \gets$ \texttt{true}
      \ForAll{$C^\prime \colon (S^\prime, C^\prime) \in pairs$}
      \State $ch\_root \gets ch\_root\; \wedge$
      \\\hspace{9.35em}\Call{Decomp}{$S^\prime, C^\prime, k$}
      \EndFor
      \State $root \gets root \vee ch\_root$
      \EndFor
      \State \Return $root$
    \EndProcedure
  \end{algorithmic}
\end{algorithm}

The algorithm depends on a dictionary $\mathcal{D}$ mapping a vertex set $S_i$ to connected components $[C_{i1} \ldots C_{ij}]$ that result from its removal from $G$. This dictionary is needed since, unlike substrings, subgraphs can be composed of each other in arbitrary ways. Given a candidate external node set, the connected components are the possible subgraphs that must be covered by child nodes in the equivalent HRG derivation tree.

Algorithm \ref{alg:decomp} recognizes $k$-width TDs of $G$. It equivalently applies an HRG whose rules have right-hand sides with no more than $k + 1$ external nodes. The term $\mathcal{N}_S(C) \subseteq S$ refers to the vertices in $S$ that subgraph $C$ is incident to. The context-free nature of hyperedge replacement is revealed by the fact that $C$ connects to the rest of $G$ only through vertices $\mathcal{N}_S(C)$---the external vertices that anchor a hypergraph to its host. This property enables a divide-and-conquer approach, where the node states can be reused by various TDs.

Algorithm \ref{alg:decomp} is initially called with $S = \varnothing$ and $C = V$. Line 2 is the base case where subgraph $C$ merged with its neighbors in $S$ is no larger than $k + 1$. In lines 4--10, the subgraph induced by $\mathcal{N}_S(C) \cup C$ is split to verify that all descendant nodes are valid roots of a $k$-width TD\@. Note that lines 4--5 iterate through all sets of children for the subgraph, yet only one $\texttt{true}$ assignment to $ch\_root$ is needed for recognition. Since we reuse this algorithm to produce a derivation forest of TDs by replacing the $(\vee, \wedge)$ operators, we keep the exhaustive search.

\begin{figure}[t]
  \centering
  \begin{tikzpicture}[uns/.style={red!70},seen/.style={blue!70}]
    \begin{scope}[every node/.style={inner sep=.1em},font=\small,
      bag/.style={inner sep=.2em,seen}]
      \node [bag] (a1) {$\{c,\boldsymbol{h},g\}$};
      \node (a2) [below left of=a1] {$\boldsymbol{y}$};
      \node (a3) [below right of=a1] {$\boldsymbol{p}$};
      \node [uns] (a4) [below right of=a2] {$t$};
      \node [bag] (b1) at (3.5,0) {$\{c,h,g\}$};
      \node [bag] (b2) at (3.5,-.9) {$\{h,\boldsymbol{y},\boldsymbol{p}\}$};
      \node (b3) [below of=b2,yshift=.5em] {$\boldsymbol{t}$};
    \end{scope}
    \begin{scope}[>=latex]
      \path [->,dashed] (a1) edge node {} (a2) edge node {} (a3);
      \draw [->,uns] (a2) -- (a4);
      \draw [->,uns] (a4) -- (a3);
      \draw[-Implies,double distance=1.5pt] (1.6,-.75) -- (2,-.75);
      \draw [seen] (b1) -- (b2);
      \path [->,dashed] ([xshift=6pt,yshift=-8pt]b2.west) edge node {} (b3);
      \path [->,dashed] (b3) edge node {} ([xshift=-6pt,yshift=-8pt]b2.east);
    \end{scope}
  \end{tikzpicture}
  \caption{A recursive step of Algorithm~\ref{alg:decomp}. Left: $S = \{c,h,g\},\ S^\prime = \{h,y,p\}$. Right: $S = \{h,y,p\},\ S^\prime = \{y,p,t\}$. Seen bags are blue and unseen edges red. Dashed edges link vertices of $S^\prime$ with those of $\mathcal{N}_S(C)$.}
  \label{fig:rec-step}
\end{figure}
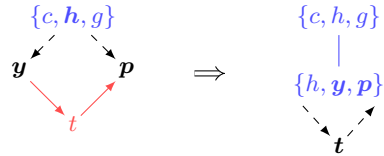

\textbf{Root-constrained search} While Algorithm \ref{alg:decomp} considers all possible TDs for a given graph, we use the rooted structure of an AMR to narrow the search. We ensure that the root bags of the extracted forest cover (i) the root vertex of an AMR and (ii) at least one edge stemming from the root. Without this constraint, the algorithm can return trees that are structurally isomorphic yet rooted at different bags. Furthermore, we wish for the forest structure to resemble that of the input AMR\@.

\textbf{Minimizing bag size} As the molecular graphs are generally larger than AMRs we further prune their TD forests by keeping derivations of minimal bag size. For each $(S, C)$ pair we store a $(k + 1)$-element bag size frequency vector, sorted in increasing order of bag size. Given multiple derivations for $(S, C)$, each derivation is assigned the sum of its $(S^\prime, C^\prime)$ frequency vectors. We only retain the derivations whose frequency vectors have the lowest lexicographical order.

\section{Graph Transformer}
The baseline system \cite{cai2020graph} for AMR-to-text generation adapts the Transformer to graph inputs. To do so it injects path embeddings between vertex pairs into encoder self-attention. For molecular property prediction we simply feed edge embeddings from the forest encoder to a baseline GCN model. We describe these changes while briefly covering the Transformer.

\subsection{Encoder}
The encoder computes embeddings of source tokens $\boldsymbol{x} = (x_1,\ldots,x_n)$ based on their initial embeddings. Given $d$-dimensional embeddings $X \in \mathbb{R}^{n \times d}$, the encoder uses dot-product attention to apply function $e\colon \mathbb{R}^{n \times d} \to \mathbb{R}^{n \times d}$. At each sub-layer, different attention heads indexed by $h$ can update positions of $\boldsymbol{x}$ in parallel:
\begin{align*}
  \mathrm{Attn}(X) &= \bigparallel_{h = 1}^H A^h (XV^h),\\
  A^h &= \sigma\left(\frac{1}{\sqrt{m}}(XQ^h)(XK^h)^\top\right),
\end{align*}
where $Q_h, K_h \in \mathbb{R}^{d \times m}$ and $V_h \in \mathbb{R}^{d \times d}$ are query, key, and value projections; $\bigparallel$ represents matrix concatenation and $\sigma$ refers to the $\softmax$ function.

Besides the $\mathrm{Attn}$ function, a sub-layer includes two linear projections and layer normalization. Residual connections between sub-layers to allow the encoder more flexibility in combining them.

\textbf{Relation encoder} The original Transformer used positional encodings of string $\boldsymbol{x}$ as added inputs to the encoder. These positional encodings relay information about input structure to the self-attention mechanism. In contrast to string tokens, however, unordered graph vertices $V = \{v_1 \ldots v_n\}$ cannot be assigned unique positions. To remedy this, \citeauthor{cai2020graph} learn relation embeddings $\widetilde{R} \in \mathbb{R}^{n \times n \times d}$ for each directed edge between vertices. These $\widetilde{R}$ are summed with $X$ in the dot-product attention between a query $i$ and at $j$: $\big((X_i + \widetilde{R}_{ij})Q\big) \big((X_j + \widetilde{R}_{ji})K\big)^\top$. Thus $\widetilde{R}_{ij}$ has a direct influence on the attention between the two indices. To learn $\widetilde{R}_{ij}$, they apply bidirectional GRUs along relation labels of the shortest path between vertices $v_i$ and $v_j$. To connect all vertex pairs, they add reverse edges and a global vertex $v_0$ that is adjacent to all vertices. Overall, leveraging graph paths in this way injects structure into the encoder.

\subsection{Decoder}
The decoder stack is analogous the encoder and applies function $d: \mathbb{R}^{n \times d} \to \mathbb{R}^{m \times d}$, where $m$ is the length of the target sequence $\boldsymbol{y}$. In addition to self-attention on the current decoded embeddings $Y \in \mathbb{R}^{(t - 1) \times d}$, it also performs cross-attention on the encoder output $X^\prime$. In the latter case, $Q$ applies to $Y$ and $K,V$ apply to $X^\prime$.

Following previous NLG models, \citeauthor{cai2020graph} employ a copy mechanism \cite{gu2016incorporating} to handle sparsity in the output distribution. The decoder can mix between generating target tokens and copying concept labels from the input AMR\@. It learns a distribution $p_{\mathrm{cp}} \in [0, 1]$ based on decoder hidden state $Y_t$:
\begin{align*}
  p(y_t | Y_t) &= ( 1 - p_{\mathrm{cp}})\hat{p}(y_t | Y_t) + p_{\mathrm{cp}} \sum_{s \in \mathcal{C}(y_t)} A_{ts},
\end{align*}
where $\hat{p}(y_t | Y_t)$ is the initial decoder target distribution, $\mathcal{C}(y_t)$ is the set of concepts corresponding to $y_t$, and $A \in m \times n$ is a cross-attention matrix between tokens and concepts.

\section{Forest Encoder}
\label{sec:forest-enc}
We use the inside-outside algorithm to encode nodes of an AMR's TD derivation forest. Forest node embeddings are used to update relations covered by their associated bags.

\subsection{Node initialization}
\label{sec:node-init}
The states of forest nodes are initialized based on the content and structure of their associated bags. Both inside and outside embeddings are initialized in this way. Edge labels in the bag's induced subgraph represent its content, while motif labels capture its structure. A motif refers to an unlabeled, canonical subgraph structure; isomorphic subgraphs map to the same motif.

Given matrix $A_s \in \mathbb{Z}^{k\times k}$ of relation labels in subgraph $s$, we embed all non-diagonal entries $R_s \in \mathbb{R}^{k(k - 1)r}$. We denote the motif embedding for subgraph $s$ by $M_s \in \mathbb{R}^m$. The bag embedding is computed as
\begin{align*}
  B_s &= \relu\left(W_1\left[R_s; M_s\right] + \mathbf{b}_1\right),
\end{align*}
where $W_1 \in \mathbb{R}^{(k(k - 1)r + m) \times h}$. Forest nodes mapping to the same bag share the same embedding, while those created to binarize the forest receive the zero vector. The forest root is encoded as a learned bias since it maps to an empty bag.

\subsection{Inside pass}
The inside algorithm iterates all ways to derive a nonterminal bottom-up. We apply this framework to compute forest node embeddings, training the model to learn derivation weights.

We first define a node's leaf-relative depth as
\begin{align}
  \label{eq:rel-depth}
  d_a &=
  \begin{cases}
    0 & \text{if }\mathcal{C}(a) = \varnothing\\
    \max_{b \in \mathcal{C}(a)} d_b + 1 & \text{otherwise},
  \end{cases}
\end{align}
where $\mathcal{C}(\cdot)$ selects the children of a given node. These values are used to find relative depths of parent-child nodes and parallelize batched updates.

Since the forest is binarized, each node has at most $2y$ children for $y$ derivations. A node embedding is a weighted sum of derivation states, which are each a function of two child node states. We denote the embedding of node $a$ by $e_a \in \mathbb{R}^h$.

\textbf{Child combination} Derivations are encoded based on the node embeddings and parent-relative depths of their children. The depth $d_a - d_b$ of child node $b$ relative to its parent $a$ is represented by a learned embedding $\ell_{d_a - d_b} \in \mathbb{R}^d$. Derivation states are found as
\begin{align*}
  \tilde{v}_b &= \left[e_b; \ell_{d_a - d_b}\right],\\
  v_r &= \relu\left(W_2\left[\tilde{v}_{b_1}; \tilde{v}_{b_2}\right] + \mathbf{b}_2\right),
\end{align*}
where $b_1, b_2$ are the children in derivation $r$ and $W_2 \in \mathbb{R}^{2(h + d) \times h}$.

\textbf{Derivation attention} To sum over derivation states, we apply Transformer cross-attention queried by parent node state $e_a$ and keyed by all derivations $v_r$ for $r \in [1, y]$.

\subsection{Outside pass}
Using results of the inside pass, we embed all outer contexts of a nonterminal. The outside value includes derivations of a forest node's ancestors rather than its descendents. An outside derivation consists of a parent's outside embedding and a sibling's inside one. Since both inside and outside passes compose a derivation from two predecessor embeddings, they can share model parameters.

As nodes are encoded top-down instead of bottom-up, we find the depth of each node relative to the forest root instead of the furthest leaf. As such, root-relative depth $\bar{d}_a$ follows eq.~\ref{eq:rel-depth} but replaces $\mathcal{C}(\cdot)$ with the set of a node's parents $\mathcal{P}(\cdot)$.

\textbf{Parent-sibling combination} Analogous to encoding an inside derivation, embeddings from a parent and sibling are combined to produce an outside derivation:
\begin{align*}
  \tilde{u}_c &=
  \begin{cases}
    \left[\bar{e}_c; L_{\bar{d}_c - \bar{d}_a}\right] & \text{if } c \in \mathcal{P}(a)\\
    \left[e_c; \mathbf{0}\right] & \text{otherwise}
  \end{cases}\\
  u_r &= \relu\left(W_3\left[\tilde{u}_{c_1}; \tilde{u}_{c_2}\right] + \mathbf{b}_3\right),
\end{align*}
where $c_1 \in \mathcal{P}(a)$ is the parent and $c_2 \in \mathcal{C}(c_1)$ the sibling in derivation $r$. If node $a$ has $z$ unique parents, then $r \in [1, z]$. The outside embedding of node $c$ is written as $\bar{e}_c$.

\textbf{Derivation attention} Neural parameters from the inside pass cross-attention module are reused to combine outside derivation states. The attention mechanism uses outside parent node state $\bar{e}_a$ as the query and derivations $u_r$ for $r \in [1, z]$ as the keys.

\subsection{Relation updates}
In order to update AMR relations as a function of forest nodes, we rely on the edge cover property of a single TD\@. Since each edge exists in exactly one TD bag, we compute an expected tree from the forest such that this property is upheld.

Recall that the inside pass computes derivation attention weights for each node and all sets of its children. This means that each pairwise arc between nodes receives a weight normalized by the sum of all arc weights for the same source node. To find the probability of each node appearing in the final parsed tree, we apply sum-product over these arc weights top-down over the forest.

More concretely, let $A$ be a node and $B_r, C_r$ its children for derivation $r \in [1, y]$. The probability $c(A)$ depends on all ways of reaching node $A$ from the root. Thus, we first initialize $c(\textsc{root}) = 1$ and must compute $c(\cdot)$ for all nodes between the root and $A$ before computing $c(A)$. If we process the rule $A \to B_r\,C_r$, the following quantity is added to each of $c(B_r)$ and $c(C_r)$:
\begin{align}
  \label{eq:forest-msg}
  m_{A \to B_r,C_r} &= c(A) \cdot w(A \to B_r\,C_r).
\end{align}
This simply extends the path prefix ending in $A$ by one rule application. The values $w(\cdot)$ are derivation attention weights from the inside pass, where $\sum_{r^\prime \in [1, y]} w(A \to B_{r^\prime}\,C_{r^\prime}) = 1$. Processing rules in the forest top-down ensures that we compute all of a node's predecessor weights before its own.

\textbf{Node aggregation} While forming an expected tree from a weighted sum of forest nodes, we update AMR relation embeddings as a function of relevant node embeddings. At each parent-child arc we take the set of relations covered by the bag associated with the child node. For example, let $X, X^\prime$ be the external vertex sets linked to nodes $A, B_r$ as described in \S\ref{sec:rel-hrg}. The set of updated relations is
\begin{align*}
  \mathcal{R}_{X, X^\prime} &= \{(u, v) \in X^\prime \colon |\{u, v\} \cap (X \cap X^\prime)| < 2\},
\end{align*}
which ignores relations whose endpoints both fall in the intersection of parent and child bags. This filtering helps avoid relations previously covered by antecedent nodes. Furthermore, we ignore cases where relations that are self-loops (i.e., $u = v$) or do not exist the AMR\@. This restricts relations to those that affect vertex self-attention. The relations $(u, v) \in \mathcal{R}_{X, X^\prime}$ are updated as
\begin{align*}
  \mathbf{r}_{u,v} \mathrel{+}= m_{A \to B_r,C_r}\cdot \relu\left(W_4\left[e_A; \bar{e}_A\right] + \mathbf{b}_4\right).
\end{align*}
The above update takes into account both inside and outside node embeddings of parent $A$, weighted by the probability of choosing derivation $r$. Updating relations in this way ensures that each relation receives a total weight of one across all derivations in the forest.

\section{Experiments}
For AMR-to-text generation the baseline system \cite{cai2020graph} computes pairwise shortest paths to encode relations for vertex self-attention. Our system replaces these paths with a forest encoder that embeds relations from forest node states, which are found via inside-outside. We use the \ldc{} sembank of 36,521 training, 1,368 development, and 1,371 test AMR-sentence pairs. Output sentences are evaluated via BLEU \cite{papineni2002bleu} and \chrf{} \cite{popovic2017chrf++}.

For molecular property prediction we built atop a GCN-based model called a directional graph network \cite[DGN]{beaini2021directional} that propagates information via directional derivatives. \mol{} contains 32,901 training, 4,113 validation, and 4,113 test molecule-label pairs. A model must predict whether a molecule inhibits HIV virus replication. The area under the receiver operating curve (ROC-AUC) is the evaluation metric.

\begin{table}[t]
  \centering
  \begin{tabular}{@{}>{\quad}l c c@{}}
    \toprule
    & BLEU & \chrf{}\\
    \midrule
    \newcite{cai2020graph} & 29.8 & 59.4\\
    \tdp{} & \textbf{30.5} & \textbf{59.7}\\
    \quad\tdpm{} & 30.0 & 59.1\\
    \quad\tree{} & 30.0 & 59.0\\
    \bottomrule
  \end{tabular}
  \caption{\ldc{} test split scores.}
  \label{tab:results}
\end{table}

\subsection{Preprocessing}
\textbf{\ldc{}} We adopt the same preprocessing pipeline as the baseline, which was first introduced by \citet{konstas2017neural}. The pipeline removes sense tags (e.g., `go-01' $\to$ `go') and variables (e.g., `b / boy' $\to$ `boy') from concept labels. The AMRs are recategorized to simplify graph structure, which involves replacing named entity subgraphs with category concepts and standardizing dates.

\subsection{Setup}
\textbf{\ldc{}} As we use motif embeddings to initialize bag embeddings (\S\ref{sec:node-init}), we ablate them in \tdpm{} to test effectiveness. While embedding bag-level relation labels $R_s$ is enough to uniquely encode bags, we assert that motifs provide added graph structure to the edges covered per bag. We also ablate the use of TD forests as opposed to single trees in the \tree{} setting.

The \tdp{} model adopts baseline hyperparameters when possible. New hyperparameters are tuned manually based on validation BLEU and \chrf{}. Vertex and token embeddings are of size 300 and randomly initialized. Their encoders include character-level CNNs with filter size 3 and projection size 256. Attention blocks have hidden states of size 512, feed-forward projections of size 1024, and 8 attention heads. Model parameters are optimized using \textit{Adam} \cite{kingma2015adam} with default settings of $\alpha = 0.001$, $\beta_1 = 0.9$, $\beta_2 = .999$, and $\eta = 10^{-8}$. Training for both datasets is done on a 16GB Tesla V100 GPU.

In the ablated \tdpm{}, the size of embedding $R_s$ is kept the same as in \tdp{} for consistency. The original $[R_s, M_s]$ sizes are $[64k(k - 1), 32]$ with $k = 4$ as the maximum number of AMR concepts per bag. The AMRs are parsed into TDs of maximum width 3, while those with higher treewidth skip the forest encoder altogether.

\textbf{\mol{}} \tdp{} adopts all hyperparameters from DGN. Vertex embeddings are size 70 and there are five convolution layers. In \tdp{} parser parameters are jointly optimized with existing ones. \textit{Adam} is used and learning rate is further halved after every twenty-epoch plateau in validation ROC-AUC. For \tdp{} $[R_s, M_s]$ sizes are $[32k(k - 1), 16]$ for $k = 4$ and edge embeddings are of size 8. We run the author-provided training script five times for each of DGN and \tdp{} in the experiments.

\begin{table}[t]
  \centering
  \begin{tabular}{@{}l c c@{}}
    \toprule
    & Test & Validation\\
    \midrule
    \citet{beaini2021directional} & 78.29 (0.35) & 84.35 (0.34)\\
    \tdp{} & \textbf{80.21 (0.73)} & \textbf{85.32 (0.30)}\\
    \bottomrule
  \end{tabular}
  \caption{\mol{} ROC-AUC percentages with standard deviation in parentheses.}
  \label{tab:results2}
\end{table}

\subsection{Results}
\textbf{\ldc{}} As shown in Table~\ref{tab:results} \tdp{} gains a 0.7 BLEU and 0.3 \chrf{} margin over \citet{cai2020graph}. \tdpm{} and \tree{} reveal the benefits of motif embeddings and parsing forests instead of trees. Both settings reduce BLEU by 0.5, while \chrf{} drops by 0.3 in \tdpm{} and 0.4 in \tree{}. These scores support the grounding effect of motif embeddings in the AMR topology. In addition, encoding forests rather than singular TDs appears to bolster the model's view of graph structure.

\textbf{\mol{}} Table~\ref{tab:results2} lists the mean ROC-AUC percentages over five trials for test and validation splits. \tdp{} offers a clear advantage over DGN, with a 1.92\% increase test ROC-AUC. During training it also reaches a higher validation ROC-AUC by 0.97\%. We note that the evaluation splits of \mol{} are based on scaffold splitting, which tries to split the data into maximally diverse subsets according to molecular structure. Thus, these results provide more evidence that the forest encoder can generalize over global graph features.

\begin{figure}[t]
  \centering
  \captionsetup[subfigure]{justification=centering}
  \subfloat[Reentrancy count.]{
    \centering
    \begin{tikzpicture}
      \centering
      \begin{axis}[name=one, width=8cm, height=5cm, legend style={font=\scriptsize}, xtick={0,1,2,3,4}, xticklabels={0,1–2,3–4,5–6,7+}, cycle list/Set1, ylabel={\small{\chrf{}}}, legend cell align={left}]
        \addplot+[only marks, mark=triangle, mark size=1.5pt] plot[error bars/.cd, y dir=both, y explicit] table [x expr={\thisrow{x}-0.25}, y=c,y error=ec,col sep=comma] {table/re_out_04-11.csv};
        \addplot+[only marks, mark=diamond, mark size=1.5pt] plot[error bars/.cd, y dir=both, y explicit] table [y=f,y error=ef,col sep=comma] {table/re_out_05-06.csv};
        \addplot+[only marks, mark=star, mark size=1.5pt] plot[error bars/.cd, y dir=both, y explicit] table [x expr={\thisrow{x}+0.25}, y=f,y error=ef,col sep=comma] {table/re_out_04-11.csv};
	\addlegendentry{\cai{}}
        \addlegendentry{\tdpm{}}
	\addlegendentry{\tdp{}}
      \end{axis}
    \end{tikzpicture}
    \label{fig:reent-res}
  }
  \captionsetup[subfigure]{justification=centering}
  \subfloat[Diameter.]{
    \centering
    \begin{tikzpicture}
      \centering
      \begin{axis}[name=one, width=8cm, height=5cm, legend style={font=\small}, xtick={0,1,2,3,4,5,6,7}, xticklabels={0,1,2,3,4,5,6,7+}, cycle list/Set1, ylabel={\small{\chrf{}}}]
        \addplot+[only marks, mark=triangle, mark size=1.5pt] plot[error bars/.cd, y dir=both, y explicit] table [x expr={\thisrow{x}-0.25}, y=c,y error=ec,col sep=comma] {table/di_out_04-11.csv};
        \addplot+[only marks, mark=diamond, mark size=1.5pt] plot[error bars/.cd, y dir=both, y explicit] table [y=f,y error=ef,col sep=comma] {table/di_out_05-06.csv};
        \addplot+[only marks, mark=star, mark size=1.5pt] plot[error bars/.cd, y dir=both, y explicit] table [x expr={\thisrow{x}+0.25}, y=f,y error=ef,col sep=comma] {table/di_out_04-11.csv};
      \end{axis}
    \end{tikzpicture}
    \label{fig:diam-res}
  }
  \captionsetup[subfigure]{justification=centering}
  \subfloat[Treewidth.]{
    \centering
    \begin{tikzpicture}
      \centering
      \begin{axis}[name=one, width=8cm, height=5cm, legend style={font=\small}, xtick={0,1,2,3,4}, xticklabels={0,1,2,3,4}, cycle list/Set1, ylabel={\small{\chrf{}}}]
        \addplot+[only marks, mark=triangle, mark size=1.5pt] plot[error bars/.cd, y dir=both, y explicit] table [x expr={\thisrow{x}-0.25}, y=c,y error=ec,col sep=comma] {table/tw_out_04-11.csv};
        \addplot+[only marks, mark=diamond, mark size=1.5pt] plot[error bars/.cd, y dir=both, y explicit] table [y=f,y error=ef,col sep=comma] {table/tw_out_05-06.csv};
        \addplot+[only marks, mark=star, mark size=1.5pt] plot[error bars/.cd, y dir=both, y explicit] table [x expr={\thisrow{x}+0.25}, y=f,y error=ef,col sep=comma] {table/tw_out_04-11.csv};
      \end{axis}
    \end{tikzpicture}
    \label{fig:tw-res}
  }
  \caption{Test split average sentence-level \chrf{} over binned graph connectivity metrics.}
\end{figure}
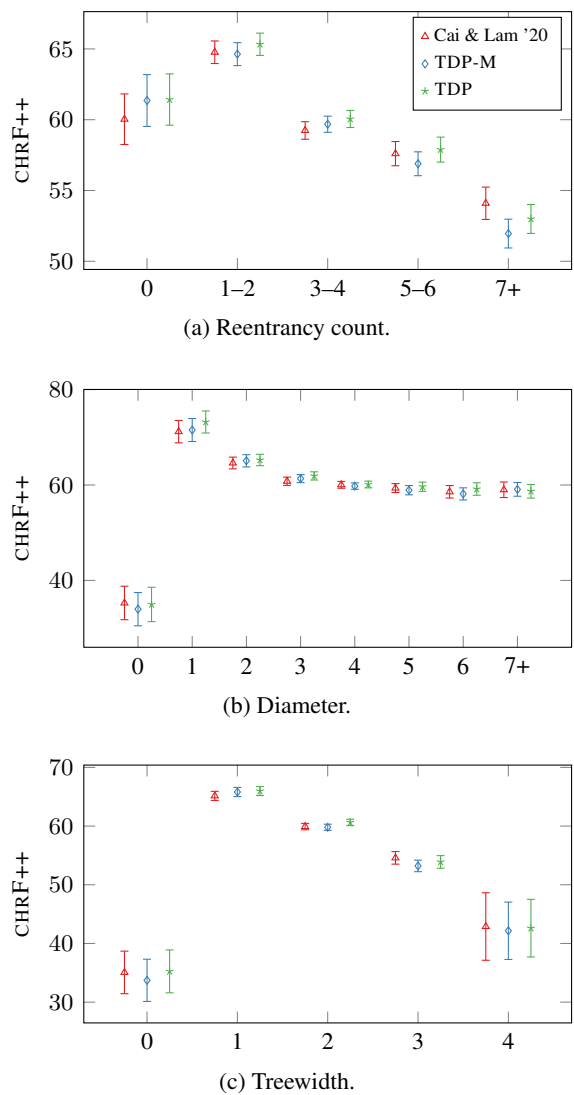

\subsection{Analysis}
Focusing on AMR-to-text generation, we proposed that TDs can help close the structural gap between graphs and strings. Specifically, TDs summarize global graph structure while their acyclicity imposes an order on the vertices. If this is valid we would expect to see better results on structurally complex graphs in particular. In this section we inspect how sentence-level \chrf{} varies with AMR reentrancy count, diameter, and treewidth. We also compare several sentences generated by the baseline and \tdp{}.

\begin{table*}[t]
  \footnotesize
  \centering
  \begin{tabularx}{\textwidth}{c l X}
    \toprule
    (1) & \rf{} & columbia is the only country that \textbf{currently has a policy} of targeting \textbf{drug trafficking aircraft} .\\
        & \cais{} & columbia is the only country with the target of \textbf{drug traffickers} .\\
        & \tdp{} & columbia is the only country that \textbf{currently has the policy} of targets for \textbf{drug trafficking aircraft} .\\
    \midrule
    (2) & \rf{} & questions have \textbf{emerged} regarding the \textbf{shipment of arms en route to kenya} and of a possible clandestine arms deal between kenya and south sudan .\\
	& \cais{} & the question with regard to kenya and south sudan a clandestine arms deal at its possible and a clandestine arms deal .\\
        & \tdp{} & the question \textbf{emerged} in regards to the \textbf{shipment of the route to kenya} and possibly a clandestine arms deal with south sudan .\\
    \midrule
    (3) & \rf{} & \textbf{if you get out ahead} of the need you are accused of \textbf{causing sprawl} .\\
        & \cais{} & \textbf{when you got ahead} of need , you are accused of \textbf{being of being the sprawl} .\\
        & \tdp{} & \textbf{if you get ahead} of the need , you are accused of \textbf{causing a sprawl} .\\
    \midrule
    (4) & \rf{} & the preliminary \textbf{competition for the women 's} 10 - meter platform \textbf{diving} will be held this afternoon .\\
        & \cais{} & this afternoon a preliminary \textbf{compete between women and} 10 - meter platform will be held !\\
        & \tdp{} & this afternoon, preliminary \textbf{competition for the women 's} 10 - meter platform \textbf{dive} is held .\\
    \bottomrule
  \end{tabularx}
  \caption{Model-generated sentences; \rf{} is the reference and \cai{} is \citet{cai2020graph}.}
  \label{tab:sample-snts}
\end{table*}

\textbf{Reentrancy} As reentrancies add cycles to an AMR, higher reentrancy counts should correlate with graph complexity. They are also of interest since TDs explicitly eliminate cycles from a graph and we hypothesize that this aligns them more closely with strings. The encoder by \citet{cai2020graph}, however, relies only on shortest pairwise paths between vertices and may be susceptible to cycles. Figure~\ref{fig:reent-res} shows that as reentrancy count increases \tdp{} generally maintains a lead over the baseline. The only exception is the last bucket of AMRs that have at least seven reentrancies. Perhaps the baseline's RNN path encoder gives it an edge on highly reentrant cases. \tdpm{} suffers a performance hit relative to \tdp{} on AMRs with over four reentrancies, which suggests that motif embeddings are especially helpful for such AMRs.

\textbf{Diameter} The maximum pairwise distance between AMR vertices is another salient metric. Among connected graphs of size $n$, a fully connected graph has diameter 1 while a linear graph has diameter $n - 1$. As such, graphs of greater connectivity generally have lower diameter. Figure~\ref{fig:diam-res} shows that \tdp{} offers the greatest boost in \chrf{} for AMRs of diameter 1. It maintains a less dramatic advantage for diameters up to 6. Clearly the forest encoder can simplify global topology across a range of graph structures.

\textbf{Treewidth} As defined in \S\ref{sec:td-def}, treewidth correlates with the connectivity of a given graph. Intuitively, a graph with large, tight clusters needs larger bags to cover all its edges in a TD. As \tdp{} directly captures higher-level AMR structure through TDs, it may be less affected by high treewidth AMRs. Figure~\ref{fig:tw-res} indicates that \tdp{} outperforms the baseline on AMRs of treewidth up to 2. The performance dip on higher treewidths may be due to data sparsity, which hurts \tdp{}'s ability to learn motif embeddings of larger bags.

\textbf{Manual inspection} We next examine differences between the models' predicted sentences. Sentences in Table~\ref{tab:sample-snts} highlight differences in how the models handle linguistic phenomena. The AMRs of (1) and (2) have reentrancy counts and diameters of 3--4, while those of (3) and (4) have 1--2 reentrancies and diameters of 3. All examples have treewidth 2 except for (4), which has treewidth 1.

In (1) the baseline drops the phrase \textit{currently has a policy} and substitutes \textit{drug trafficking aircraft} with \textit{drug traffickers}. These omittances suggest that the baseline's encoder is not considering global structure, as \tdp{} does not make the same errors. The reference sentence of (2) contains several clauses and long-range dependencies between the verb and noun phrases. The baseline omits the verb \textit{emerged} and exhibits a lack of syntactic structure in the second half of the sentence. \tdp{} maintains integrity of the two clauses referring to a \textit{shipment of arms} and \textit{arms deal}, though it drops the word \textit{arms} in the former. Example (3) involves causality that the baseline fails to express; instead of \textit{if you get out ahead} it chooses \textit{when you got ahead}, which no longer describes a hypothetical scenario. It repeats itself at the sentence's end, replacing \textit{causing} with \textit{being}. Lastly, (4) reveals the importance of preserving relations between concepts. Whereas \tdp{} correctly refers to a \textit{women 's} diving competition, the baseline describes a \textit{compete between women and \ldots platform}. It unfortunately makes no mention of \textit{diving}. This and previous examples illustrate how inaccuracies in the graph encoder may trickle down to sentence decoding.

\section{Conclusion}
We presented a way to parse TD forests within a graph encoder to bias it on global graph structure. With the structural mismatch between AMRs and their target sentences, TDs balance traits of both. Due to their equivalence to HRGs during parsing, TDs also serve as a language over source graphs. Our forest encoder efficiently forms TD node embeddings using the inside-outside algorithm. Experiments on AMR-to-text generation validate the use of motif embeddings and encoding TD forests rather than single trees. In addition to higher corpus-level BLEU and \chrf{} scores, our models succeed on structurally complex AMRs. We also reveal the impact of our forest encoder on coherence of decoded sentences. Finally, promising results on molecular property prediction support the generality of our parser to other model architectures and data domains. We leave further exploration in this direction for future work.

\paragraph{Acknowledgments} This work was supported by NSF awards IIS-1813823 and CCF-1934962.

\bibliographystyle{acl_natbib}
\bibliography{acl2021}

\begin{thebibliography}{19}
\expandafter\ifx\csname natexlab\endcsname\relax\def\natexlab#1{#1}\fi

\bibitem[{Arnborg et~al.(1991)Arnborg, Lagergren, and Seese}]{arnborg1991easy}
Stefan Arnborg, Jens Lagergren, and Detlef Seese. 1991.
\newblock \href {https://doi.org/10.1016/0196-6774(91)90006-K} {Easy problems
  for tree-decomposable graphs}.
\newblock \emph{Journal of Algorithms}, 12(2):308--340.

\bibitem[{Banarescu et~al.(2013)Banarescu, Bonial, Cai, Georgescu, Griffitt,
  Hermjakob, Knight, Koehn, Palmer, and Schneider}]{banarescu2013abstract}
Laura Banarescu, Claire Bonial, Shu Cai, Madalina Georgescu, Kira Griffitt, Ulf
  Hermjakob, Kevin Knight, Philipp Koehn, Martha Palmer, and Nathan Schneider.
  2013.
\newblock Abstract {M}eaning {R}epresentation for sembanking.
\newblock In \emph{Proceedings of the 7th Linguistic Annotation Workshop and
  Interoperability with Discourse}, pages 178--186.

\bibitem[{Beaini et~al.(2021)Beaini, Passaro, Létourneau, Hamilton, Corso, and
  Liò}]{beaini2021directional}
Dominique Beaini, Saro Passaro, Vincent Létourneau, William~L. Hamilton,
  Gabriele Corso, and Pietro Liò. 2021.
\newblock \href {http://arxiv.org/abs/2010.02863} {Directional graph networks}.

\bibitem[{Beck et~al.(2018)Beck, Haffari, and Cohn}]{beck2018graph}
Daniel Beck, Gholamreza Haffari, and Trevor Cohn. 2018.
\newblock Graph-to-sequence learning using gated graph neural networks.
\newblock In \emph{Proceedings of the 56th Annual Meeting of the Association
  for Computational Linguistics (ACL-18)}, pages 273--283.

\bibitem[{Cai and Lam(2020)}]{cai2020graph}
Deng Cai and Wai Lam. 2020.
\newblock Graph {T}ransformer for graph-to-sequence learning.
\newblock In \emph{Proceedings of the 34th AAAI Conference on Artificial
  Intelligence (AAAI-20)}.

\bibitem[{Gu et~al.(2016)Gu, Lu, Li, and Li}]{gu2016incorporating}
Jiatao Gu, Zhengdong Lu, Hang Li, and Victor O.~K. Li. 2016.
\newblock Incorporating copying mechanism in sequence-to-sequence learning.
\newblock In \emph{Proceedings of the 54th Annual Meeting of the Association
  for Computational Linguistics (ACL-16)}, pages 1631--1640.

\bibitem[{Hu et~al.(2020)Hu, Fey, Zitnik, Dong, Ren, Liu, Catasta, and
  Leskovec}]{hu2020ogb}
Weihua Hu, Matthias Fey, Marinka Zitnik, Yuxiao Dong, Hongyu Ren, Bowen Liu,
  Michele Catasta, and Jure Leskovec. 2020.
\newblock Open graph benchmark: Datasets for machine learning on graphs.
\newblock \emph{arXiv preprint arXiv:2005.00687}.

\bibitem[{Jin et~al.(2018)Jin, Barzilay, and Jaakkola}]{jin2018junction}
Wengong Jin, Regina Barzilay, and Tommi Jaakkola. 2018.
\newblock Junction tree variational autoencoder for molecular graph generation.
\newblock In \emph{International Conference on Machine Learning}, pages
  2323--2332. PMLR.

\bibitem[{Jones et~al.(2013)Jones, Goldwater, and Johnson}]{jones2013modeling}
Bevan Jones, Sharon Goldwater, and Mark Johnson. 2013.
\newblock Modeling graph languages with grammars extracted via tree
  decompositions.
\newblock In \emph{Proceedings of the 11th International Conference on Finite
  State Methods and Natural Language Processing}, pages 54--62.

\bibitem[{Kingma and Ba(2015)}]{kingma2015adam}
Diederik~P. Kingma and Jimmy Ba. 2015.
\newblock Adam: A method for stochastic optimization.
\newblock In \emph{Proceedings of the 3rd International Conference on Learning
  Representations (ICLR-15)}.

\bibitem[{Konstas et~al.(2017)Konstas, Iyer, Yatskar, Choi, and
  Zettlemoyer}]{konstas2017neural}
Ioannis Konstas, Srinivasan Iyer, Mark Yatskar, Yejin Choi, and Luke
  Zettlemoyer. 2017.
\newblock Neural {AMR}: Sequence-to-sequence models for parsing and generation.
\newblock In \emph{Proceedings of the 55th Annual Meeting of the Association
  for Computational Linguistics (ACL-17)}, pages 146--157.

\bibitem[{Lautemann(1988)}]{lautemann1988decomposition}
Clemens Lautemann. 1988.
\newblock Decomposition trees: structured graph representation and efficient
  algorithms.
\newblock In \emph{Colloquium on Trees in Algebra and Programming}, pages
  28--39. Springer.

\bibitem[{Papineni et~al.(2002)Papineni, Roukos, Ward, and
  Zhu}]{papineni2002bleu}
Kishore Papineni, Salim Roukos, Todd Ward, and Wei-Jing Zhu. 2002.
\newblock {BLEU}: a method for automatic evaluation of machine translation.
\newblock In \emph{Proceedings of the 40th Annual Meeting of the Association
  for Computational Linguistics (ACL-02)}, pages 311--318.

\bibitem[{Popovi{\'c}(2017)}]{popovic2017chrf++}
Maja Popovi{\'c}. 2017.
\newblock chr{F}++: words helping character n-grams.
\newblock In \emph{Proceedings of the 2nd Conference on Machine Translation},
  pages 612--618.

\bibitem[{Song et~al.(2017)Song, Peng, Zhang, Wang, and Gildea}]{song2017amr}
Linfeng Song, Xiaochang Peng, Yue Zhang, Zhiguo Wang, and Daniel Gildea. 2017.
\newblock {AMR}-to-text generation with synchronous node replacement grammar.
\newblock In \emph{Proceedings of the 55th Annual Meeting of the Association
  for Computational Linguistics (ACL-17)}, pages 7--13.

\bibitem[{Vaswani et~al.(2017)Vaswani, Shazeer, Parmar, Uszkoreit, Jones,
  Gomez, Kaiser, and Polosukhin}]{vaswani2017attention}
Ashish Vaswani, Noam Shazeer, Niki Parmar, Jakob Uszkoreit, Llion Jones,
  Aidan~N Gomez, {\L}ukasz Kaiser, and Illia Polosukhin. 2017.
\newblock Attention is all you need.
\newblock In \emph{Advances in Neural Information Processing Systems
  (NIPS-17)}, pages 5998--6008.

\bibitem[{Wang et~al.(2020)Wang, Wan, and Jin}]{wang2020amr}
Tianming Wang, Xiaojun Wan, and Hanqi Jin. 2020.
\newblock {AMR}-to-text generation with {G}raph {T}ransformer.
\newblock \emph{Transactions of the Association for Computational Linguistics},
  8:19--33.

\bibitem[{Zhao et~al.(2020)Zhao, Walker, and Chaturvedi}]{zhao2020bridging}
Chao Zhao, Marilyn Walker, and Snigdha Chaturvedi. 2020.
\newblock Bridging the structural gap between encoding and decoding for
  data-to-text generation.
\newblock In \emph{Proceedings of the 58th Annual Meeting of the Association
  for Computational Linguistics}, pages 2481--2491.

\bibitem[{Zhu et~al.(2019)Zhu, Li, Zhu, Qian, Zhang, and
  Zhou}]{zhu2019modeling}
Jie Zhu, Junhui Li, Muhua Zhu, Longhua Qian, Min Zhang, and Guodong Zhou. 2019.
\newblock Modeling graph structure in {T}ransformer for better {AMR}-to-text
  generation.
\newblock In \emph{Proceedings of the 2019 Conference on Empirical Methods in
  Natural Language Processing and the 9th International Joint Conference on
  Natural Language Processing (EMNLP-IJCNLP-19)}, pages 5462--5471.

\end{thebibliography}

\end{document}